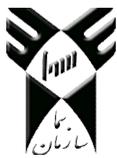
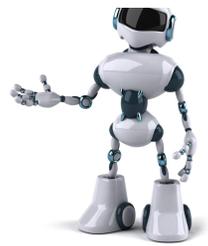

# Improving high-pass fusion method using wavelets

Hamid Reza Shahdoosti

# بهبود روش فیلتر بالا گذر در ادغام تصاویر با استفاده از موجک‌ها

حمید رضا شاه‌دوستی[1]

[1] استادیار، دانشگاه صنعتی همدان؛ *h.doosti@hut.ac.ir*

**Abstract**

In an appropriate image fusion method, spatial information of the panchromatic image is injected into the multispectral images such that the spectral information is not distorted. The high-pass modulation method is a successful method in image fusion. However, the main drawback of this method is that this technique uses the boxcar filter to extract the high frequency information of the panchromatic image. Using the boxcar filter introduces the ringing effect into the fused image. To cope with this problem, we use the wavelet transform instead of boxcar filters. Then, the results of the proposed method and those of other methods such as, Brovey, IHS, and PCA ones are compared. Experiments show the superiority of the proposed method in terms of correlation coefficient and mutual information.

**چکیده**

در یک روش مناسب ادغام تصاویر، اطلاعات مکانی تصویر تک‌رنگ طوری به تصاویر چندطیفی تزریق می‌شود که اطلاعات طیفی آنها را تخریب نکند. روش مدولاسیون بالاگذر یکی از روش‌های موفق در ادغام تصاویر است که دارای معایبی است. از جمله اینکه در این روش، از فیلترهای نوع **Box car** به منظور استخراج اطلاعات فرکانس بالای تصاویر تک‌رنگ (اطلاعات مکانی) و تزریق آنها به تصاویر چندطیفی استفاده شده است. اما ریپل‌های موجود در این نوع فیلترها، باعث پدیده حلقه حلقه شدن در تصویر ادغام شده می‌شود. در این مقاله از تبدیل موجک به منظور بهبود روش مدولاسیون بالاگذر استفاده شده است. سپس نتایج بدست آمده توسط روش پیشنهادی و سایر روش‌های موجود مانند براوی، **IHS** و روش مولفه اصلی، مقایسه شده است. کارایی روش پیشنهادی با استفاده از معیارهای مختلفی چون ضریب همبستگی و اطلاعات متقابل نشان داده خواهد شد.

**واژه‌های کلیدی**: ادغام تصاویر، مدولاسیون بالاگذر، موجک، تصویر چندطیفی.

## 1- مقدمه

توسعه و کاربرد فناوری ماهواره‌ای روز به روز در حال افزایش است و بر تنوع ماهواره‌های مشاهده گر زمین افزوده می‌شود. داده های تصویری از سطح زمین توسط سنسورهای این ماهواره ها جمع آوری می‌شوند. این سنسورها دارای دقت‌های طیفی، مکانی و رادیومتری و زمانی متنوعی هستند. بنابراین نیاز مبرمی به توسعه و گسترش روش‌ها و الگوریتم‌های ادغام اطلاعات وجود دارد تا از این امکانات برای کاربردهای متفاوت، بهره گیری مطلوب صورت پذیرد.

امکان اخذ تصاویری که هم دقت طیفی و هم دقت مکانی بالایی داشته باشند، بدلیل محدودیتهای عملی از جمله محدودیت سیگنال به نویز امکان‌پذیر نیست. ادغام تصاویر ماهواره ای یکی از روش‌هایی است که می تواند از اطلاعات مکمل تصاویر استفاده نماید به طوری‌که اطلاعات مکانی را از تصاویر با قدرت تفکیک مکانی بالا (تصاویر تک‌رنگ) و اطلاعات طیفی را از تصاویر با قدرت تفکیک بالا (تصاویر چندطیفی) با هم ادغام نموده و تصویری با دقت مکانی و طیفی بالا ایجاد می نماید.

بسته به کاربردهای مختلف و نوع زمین مورد مطالعه [1]، الگوریتم‌های گوناگونی به منظور ادغام تصاویر ارائه شده اند که از جمله انها می توان به الگوریتم براوی، IHS، PCA و موجک اشاره کرد. هر یک از این الگوریتم‌ها دارای نقاط قوت و ضعفی هستند که کمک ترکیب روش‌های مختلف می‌توان از نقطه قوت هر یک از روش‌ها استفاده نمود.

الگوریتم براوی یکی از ابتدایی ترین روش‌های پیشنهادی در ادغام تصاویر بوده است اما اضافه شدن اطلاعات مکانی در این روش منجر به تخریب اطلاعات طیفی می‌شود [2].

الگوریتم‌های IHS و PCA [3]-[4] از الگوریتم‌های مطرح در ادغام تصاویر هستند که اطلاعات مکانی را به خوبی حفظ می‌کنند اما اطلاعات طیفی در این روش‌ها نیز به خوبی حفظ نمی‌شود و این موضوع باعث به هم ریختگی رنگ‌ها در تصاویر ادغام شده می شود.

یکی دیگر از الگوریتم‌های نسبتا ساده اما کارا، الگوریتم مدولاسیون بالاگذر[2] است که اطلاعات فرکانس بالای تصویر تک‌رنگ را با یک ضریب وارد اطلاعات فرکانس بالای تصاویر چند طیفی می‌کند. این الگوریتم به منظور استخراج اطلاعات فرکانس بالای تصویر تک‌رنگ، از فیلتر های Box car استفاده می‌کند[2]. این فیلتر ها منجر به نتایج خوبی نمی شوند زیرا در حوزه فرکانس دارای ریپل‌های زیادی هستند. این روش با استفاده از مدلسازی ریاضی سنسورهای ماهواره ای پیشنهاد شده است[2].

با معرفی موجک‌ها، نتایج ادغام تصاویر به طور قابل توجهی بهبود پیدا کردند و نتایج ادغام تصاویر تا حد زیادی به ادغام ایده‌ال (تصویر مرجع) نزدیک شد. موجک‌هایی که در حال حاضر برای ادغام به کار برده می شوند دو دسته اند. 1- موجک‌های با کاهش بعد [5] و 2- موجک‌های بدون کاهش بعد [6].

نتایج ادغام نشان داده اند که به کارگیری موجک‌های با کاهش بعد (مانند موجک مالات) چندان برای ادغام تصاویر مناسب نیست زیرا منجر به ایجاد لبه‌های مصنوعی می شوند [7]. اما موجک‌های بدون کاهش بعد (مانند موجک بدون کاهش بعد Atrous) از مطرح ترین روش‌های ادغام تصاویر هستند.

از انجایی که الگوریتم مدولاسیون بالاگذر با استفاده از مدلسازی ریاضی سنسورهای ماهواره‌ای به دست آمده است و الگوریتم کارایی به نظر می‌رسد، در این مقاله قصد داریم این الگوریتم را بهبود ببخشیم و به جای استفاده از فیلترهای Box car، از موجک‌ها جهت استخراج فرکانس بالای تصویر تک‌رنگ استفاده کنیم. در بخش دوم مقاله، الگوریتم مدولاسیون بالاگذر مرور خواهد شد. در بخش سوم

مقاله، الگوریتم پیشنهادی ارائه می‌شود. بخش چهارم به ارزیابی نتایج اختصاص داده شده است و در بخش پنجم نتیجه‌گیری حاصل از مقاله ارائه خواهد شد.

## 2- مدولاسیون بالاگذر

در روش مدولاسیون بالاگذر، اطلاعات فرکانس بالای تصویر تک‌رنگ را با استفاده از یک ضریب مدولاسیون به تصاویر چند طیفی تزریق می کنیم [2]. که این ضریب برابر با تقسیم تصویر چند طیفی به تصویر تک‌رنگ پایین گذر است. رابطه ادغام با استفاده از این روش به صورت زیر است:

$$F_i = MS_i + (PAN - PAN_L)\frac{MS_i}{PAN_L} \qquad (1)$$

که در رابطه (1)، $MS_i$ و $F_i$ تصویر چند طیفی اولیه و تصویر ادغام شده باند iام هستند. $PAN$ بیانگر تصویر تک‌رنگ اولیه است. و $PAN_L$ از رابطه زیر به دست می اید:

$$PAN_L = PAN * h_0 \qquad (2)$$

که فیلتر $h_0$ یک فیلتر پایین گذر از نوع Box car است[2].

همانطور که مشخص است، حاصل $(PAN - PAN_L)$ اطلاعات فرکانس بالای تصویر تک‌رنگ را استخراج می کند. اما متاسفانه فیلتر های Box car برای استخراج اطلاعات فرکانس بالا چندان مناسب نیستند و ریپل‌های زیادی که در حوزه فرکانس ایجاد می‌کنند، منجر به پدیده حلقه شدن در تصویر ادغام شده می‌شود.

در این مقاله سعی داریم از موجک‌ها جهت استخراج اطلاعات فرکانس بالای تصویر تک‌رنگ استفاده کنیم. فیلترهای مورد استفاده در موجک‌ها از مناسبترین فیلترها هستند که ریپل در حوزه فرکانس را به حداقل می‌رسانند. در قسمت بعد اصلاح روش مدولاسیون بالاگذر با استفاده از موجک بیان شده است.

## 3- روش پیشنهادی

همانطور که بیان شد، موجک‌ها قادرند تصاویر را به باندهای مجزای فرکانسی به خوبی تجزیه کنند. موجک‌های بدون کاهش بعد (Atrous) که به منظور ادغام تصاویر به کار می روند، با استفاده از تجزیه تصویر تک‌رنگ به صفحات موجک با استفاده از تابع B_3- Cubic Spline اطلاعات فرکانس بالای تصویر تک‌رنگ را استخراج کرده و به تصویر چندطیفی تزریق می‌کنند:

$$F_i = MS_i + \sum_{j=1}^{n} w_j \qquad (3)$$

که در رابطه بالا، $w_j$ صفحات فرکانس بالای موجک هستند [6]. الگوریتم مدولاسیون بالاگذر پیشنهادی با استفاده از موجک Atrous به صورت زیر است:

$$F_i = MS_i + (\sum_{j=1}^{n} w_j) \frac{MS_i}{(PAN - \sum_{j=1}^{n} w_j)} \quad (4)$$

رابطه شماره 4 پیشنهادی به منظور ادغام تصاویر، همان رابطه مدولاسیون بالا گذر (رابطه 2) است که در آن به جای استخراج اطلاعات فرکانس بالای تصاویر تک‌رنگ با استفاده از فیلتر Box car از صفحات موجک استفاده شده و به دلیل فیلترهای مناسب در موجک‌ها، مشکل ریپل در حوزه فرکانس در این روش حل شده است. در مخرج این رابطه نیز تفاضل تصویر تک‌رنگ از صفحات فرکانس بالای خود قرار داده شده است که حاصل این مقدار همان $PAN_L$ است ولی در استخراج $PAN_L$ این بار از موجک‌ها استفاده شده است.

## 4- نتایج بدست آمده

تصاویر ماهواره IKONOS با تصویر چند طیفی با دقت مکانی 4 متر و تصویر تک‌رنگ با دقت مکانی 1 متر برای ارزیابی نتایج این مقاله استفاده شده است. ارزیابی‌ها در ادغام تصاویر معمولا به دو روش چشمی و عددی انجام می شود. روش‌های دیگری که به منظور مقایسه در این مقاله آورده شده اند عبارتند از: براوی،مدولاسیون بالا گذر، IHS، PCA و Atrous.

نتایج حاصل از پیاده سازی در شکل (1) اورده شده‌اند.

در ابتدا تصاویر را با کمک چشم تحلیل می‌کنیم. همانطور که در شکل (2- ت) دیده می شود، ادغام با استفاده از براوی موجب شده است که رنگ‌های تصویر چند طیفی اولیه به هم بریزند اما اطلاعات مکانی تصویر ادغام شده در این روش مناسب است. شکل (2- ث) نتیجه ادغام با استفاده از مدولاسیون بالاگذر را نشان می دهد. همانطور که در این شکل نمایش داده شده است، اطلاعات مکانی تصویر تک‌رنگ در این روش به خوبی تزریق نشده است. روش Atrous اطلاعات مکانی را به درستی تزریق نکرده است و این موضوع باعث شده است که تصویر ادغام شده کمی تار به نظر برسد(شکل (2- ج) را ببینید). نتایج ادغام با استفاده از روش IHS و PCA در شکل (2- چ) و (2- ه) نمایش داده شده‌اند. این روش‌ها اطلاعات مکانی تصویر تک‌رنگ را به خوبی حفظ کرده‌اند اما اطلاعات طیفی تصویر چند طیفی اولیه تخریب شده است (در مقالات دیگر نیز به وضوح به این موضوع اشاره شده است [2]). تصویر ادغام شده با استفاده از روش پیشنهادی مناسب به نظر می رسد زیرا هم اطلاعات مکانی تصویر تک‌رنگ را به خوبی حفظ کرده است و هم رنگ‌های این روش شباهت زیادی به رنگ‌های تصویر چندطیفی دارد.

ارزیابی عددی به عنوان روشی معتبر در ارزیابی روش‌های ادغام به کار می رود. معیارهای گوناگونی به منظور ارزیابی عددی تصاویر ادغام شده به کار می‌روند که در این مقاله از سه روش معتبر همبستگی خطی[6]، اطلاعات متقابل[8] و QNR [9] استفاده می کنیم.

### 4-1 همبستگی خطی)

این نوع همبستگی، مقدار وابستگی خطی بین دو متغیر را آشکار می‌کند. همانطور که می‌دانیم مقدار بیشینه وابستگی خطی بین دو متغیر برابر با 1 است. مقدار همبستگی خطی بین دو متغیر $A$ و $B$ از رابطه زیر قابل محاسبه است [6]:

$$Corr(A,B) = \frac{\sum_{j=1}^{n}(A_j - m_A)(B_j - m_B)}{\sqrt{\sum_{j=1}^{n}(A_j - m_A)\sum_{j=1}^{n}(B_j - m_B)}} \quad (5)$$

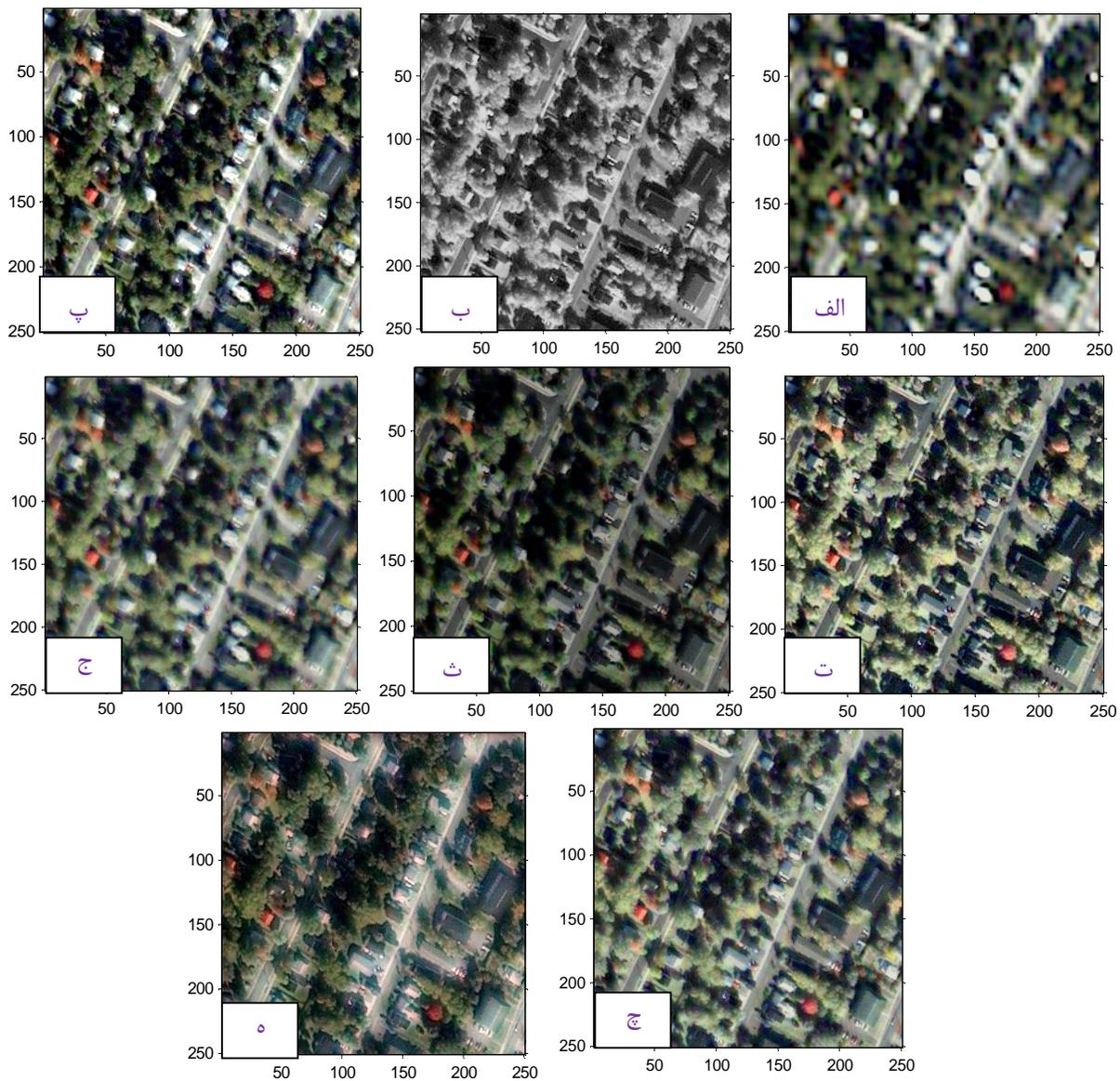

شکل (1) (الف) تصویر چند طیفی اولیه. (ب) تصویر تک‌رنگ. (پ) روش پیشنهادی. (ت) روش براوی. (ث) روش مدولاسیون بالاگذر ج) روش موجک. (چ) روش PCA. ه) روش IHS

جدول 1- ارزیابی روشهای ادغام تصاویر

| روشها | روش پیشنهادی | مدولاسیون بالاگذر | موجک | براوی | PCA | IHS |
|---|---|---|---|---|---|---|
| همبستگی خطی | **0.97** | 0.88 | 0.89 | 0.81 | 0.82 | 0.82 |
| اطلاعات متقابل | **0.54** | 0.31 | 0.36 | 0.32 | 0.33 | 0.33 |
| معیار QNR | **0.95** | 0.81 | 0.86 | 0.75 | 0.78 | 0.77 |

که در رابطه (5) $n$ نشان دهنده تعداد پیکسلها و $m$ میانگین هستند.

برای ارزیابی با استفاده از این معیار، یک بار همبستگی را بین تصویر ادغام شده و تصویر تک رنگ محاسبه می کنیم، و یک بار همبستگی را بین تصویر ادغام شده و تصویر چند طیفی اولیه محاسبه می کنیم. میانگین این دو مقدار می تواند به عنوان معیاری جهت حفظ اطلاعات طیفی و مکانی تصویر چند طیفی و تک‌رنگ اولیه به کار رود.

4-2) اطلاعات متقابل

معیار دیگری که به منظور ارزیابی ادغام تصاویر استفاده می شود، بر اساس اطلاعات متقابل است [8]. اطلاعات متقابل هرگونه وابستگی (حتی از نوع غیر خطی انرا) بین دو متغیر آشکار می‌کند. اطلاعات متقابل بین دو تصویر A و B به صورت زیر است:

$$MI(A,B) = \sum_{x,y} P_{AB}(x,y) \log \frac{P_{AB}(x,y)}{P_A(x)P_B(y)} \quad (6)$$

که در رابطه (6) $P_A$ و $P_B$ به ترتیب توزیع آماری تصاویر تک‌رنگ و چند طیفی اولیه و $P_{AB}$ توزیع آماری توأم تصاویر چند طیفی و تک رنگ اولیه است.

برای ارزیابی با استفاده از این معیار، یک بار اطلاعات متقابل را بین تصویر ادغام شده و تصویر تک رنگ اولیه محاسبه می کنیم، و یک بار این معیار را بین تصویر ادغام شده و تصویر چند طیفی اولیه محاسبه می‌کنیم. میانگین این دو مقدار به عنوان معیار اطلاعات متقابل در ارزیابی تصاویر ادغام شده به کار می‌رود [8].

4-3) معیار UIQI

معیار UIQI معیار جدید دیگری است که می توان با کمک آن به ارزیابی تصاویر ادغام شده پرداخت [9]. این معیار قادر است به همریختگی طیفی و مکانی را مدل کند. و هنگامی که تصویر ادغام شده حاوی همه اطلاعات تصویر تک رنگ و چند طیفی اولیه باشد، این مقدار برابر با 1 خواهد بود (بهترین مقدار برای این کمیت برابر با 1 است).

جدول (1) بیانگر ارزیابی عددی تصاویر ادغام شده در شکل (1) است. این جدول نشان می‌دهد که ادغام با الگوریتم اصلاح شده مدولاسیون بالاگذر با استفاده از موجک که در این مقاله پیشنهاد شده است، بهترین نتیجه را کسب کرده است.

## 5- نتایج

در این مقاله با بهره‌گیری از تبدیل موجک Atrous، روش مدولاسیون بالاگذر در ادغام تصاویر را اصلاح کردیم. در این روش، بعد از استخراج اطلاعات فرکانس بالای تصویر تک‌رنگ با استفاده از موجک، آن را با استفاده از ضریب مدولاسیون به جای اطلاعات فرکانس بالای تصویر چند طیفی قرار می‌دهیم. همانطور که در مقاله اشاره شد، این ضریب مدولاسیون از مدلسازی ریاضی سنسورهای اخذ تصویر به دست آمده است. در واقع با این کار، به جای فیلتر Box car از موجک‌ها استفاده کرده‌ایم. در ادامه، تصاویر به دست آمده را ارزیابی چشمی کردیم. اگر چه ارزیابی چشمی دقیق نیست، ولی به همریختگی تصاویر ادغام شده در برخی از روش‌ها مشهود بود. سپس تصاویر به دست امده را ارزیابی عددی کردیم. جدول‌های ارزیابی نتایج ارزیابی چشمی را تایید کردند.